\documentclass{article}
\usepackage{ijcai20}

\usepackage{times}
\usepackage{soul}
\usepackage{url}
\usepackage[hidelinks]{hyperref}
\usepackage[utf8]{inputenc}
\usepackage[small]{caption}
\usepackage{graphicx}
\usepackage{amsmath}
\usepackage{amssymb}
\usepackage{amsthm}
\usepackage{booktabs}
\usepackage{algorithm}
\usepackage{algorithmic}
\usepackage{array}
\usepackage{CJKutf8}
\newcommand*{\Ja}[1]{\begin{CJK}{UTF8}{ipxm}#1\end{CJK}}

\title{S-APIR: News-based Business Sentiment Index}

\author{
Kazuhiro Seki$^{1,3}$\footnote{Contact Author}\And
Yusuke Ikuta$^{2,3}$\\
\affiliations
$^1$Faculty of Intelligence and Informatics, Konan University\\
$^2$Faculty of Business Management, Osaka Sangyo University\\
$^3$Asia Pacific Institute of Research\\
\emails
seki@konan-u.ac.jp
}

\begin{document}

\maketitle

\begin{abstract}
This paper describes our work on developing a new business
  sentiment index using daily newspaper articles.  We adopt a
  recurrent neural network (RNN) with Gated Recurrent Units to predict
  the business sentiment of a given text.  An RNN is initially trained
  on Economy Watchers Survey and then fine-tuned on news texts for
  domain adaptation.  Also, a one-class support vector machine is
  applied to filter out texts deemed irrelevant to business
  sentiment.  Moreover, we propose a simple approach to temporally
  analyzing how much and when any given factor influences the
  predicted business sentiment.  The validity and utility of the
  proposed approaches are empirically demonstrated through a series of
  experiments on Nikkei Newspaper articles published from 2013 to
  2018.
\end{abstract}

\section{Introduction}
\label{sec:intro}

There exist business sentiment indices computed through surveys, such
as Economy Watchers
Survey\footnote{\url{https://www5.cao.go.jp/keizai3/watcher-e/index-e.html}}
and Short‐term Economic Survey of Principal Enterprise in
Japan\footnote{\url{https://www.boj.or.jp/en/statistics/tk/long_syu/index.htm/}}
in the case of Japan.  These diffusion indices (DI) play a crucial
role in decision making for governmental/monetary policies, industrial
production planning, and institutional/private investment. However,
these DI's rely on traditional surveys, which is costly and
time-consuming to conduct.

For example, Economy Watchers Survey is carried out in 12 regions of
Japan, where 2,050 preselected respondents who can observe the
regional business/economic conditions (e.g., store owners and taxi
drivers) fill out a questionnaire and then an investigative organization
in each region aggregates the surveys and calculates a DI.  As the
survey and subsequent processing take time, the DI is published only
monthly.

On the other hand, so-called alternative data, including merchandise
sales, news, microblogs, query logs, GPS location information, and
satellite images, are constantly generated and accumulated.  The
availability of such data has accelerated the development of
data-driven machine learning techniques represented by deep learning.
In econometrics, there is a growing interest in future/current
forecasts of economic and financial indices by using such alternative,
large-scale data instead of traditional
surveys~\cite{kapetanios18:_big_data_macroec_nowcas}. For example,
point of sales (POS) data were used for estimating consumer price
index (CPI)~\cite{watanabe14:_estim_daily_inflat_using_scann_data},
financial and economic reports for business
sentiment~\cite{yamamoto16eng}, newspaper for grain market prices,
stock prices, and economic
indices~\cite{chakraborty16:_predic_socio_econom_indic_using_news_event,shapiro17:_measur_news_sentim,yoshihara14:_predic_stock_market_trend_recur,yoshihara16:_lever}
and social media for stock
prices~\cite{bollen11:_twitt,levenberg14:_predic_econom_indic_web_text}.

This work focuses on textual data and uses daily newspaper articles to
develop a new business sentiment index, named S-APIR index.  In
addition, using the computed index, we propose an approach to
temporally analyzing any given factors that may influence 
the business sentiment index.


\section{Related Work}
\label{sec:related_work}

In the economic and financial domains, there are abundant textual
data, such as newspaper articles and financial reports in addition to
many numerical data. These texts are intended to be read by people,
who consider other sources of information and make decisions on
investment, financial policies, and so on. However, it is difficult
even for experts to read, grasp, and synthesize all the available
information in a limited time. Therefore, there has been much
research on computing economical/financial indices from textual
data. In the following, we summarize the representative work in
business sentiment prediction, which is the main theme of the present
work.
 


\subsection{Business Sentiment Prediction}

Economy Watchers Survey introduced in Section~\ref{sec:intro}
publishes not only the business sentiment index (hereafter called EWDI
for short) but also individual survey responses on which EWDI is
based. The survey responses contain a pair of an economic condition on
a five-point scale and a statement of the reasons why the respondent
chose the particular economic condition in natural language.  Some
example responses are shown in Table~\ref{tab:keiki_watcher}.

\begin{table*}[tb]
  \small 
  \centering
  \caption{Example responses from Economy Watchers Survey.}
  \label{tab:keiki_watcher}
  \smallskip
  \begin{tabular}{>{\centering\arraybackslash} m{12mm}
    >{\centering\arraybackslash} m{43mm} >{\centering\arraybackslash}
    m{12mm} m{80mm}}
    \toprule 
    \multicolumn{1}{c}{Region} &
                                 \multicolumn{1}{c}{Occupation}&Economic condition
    & \multicolumn{1}{c}{Statement of reasons}\\
    \midrule
    Hokkaido & Taxi driver & \multicolumn{1}{c}{$\times$} &  Although sales are declining,
                                        seasonal factors and the
                                        downturn in the economy are
                                        also affecting.\\
    North Kanto & Transportation machinery and equipment manufacturing
                                       & \multicolumn{1}{c}{$\circledcirc$} & Automobile exports to the United States are increasing.\\
    \bottomrule 
  \end{tabular}
\end{table*}

Based on the responses, EWDI is computed by first computing the
composition ratios of the five economic conditions and then taking
their weighted sum. EWDI ranges from 0 to 100 with 50 being the middle
(meaning that economic condition is neither positive or negative).

Yamamoto et al.~\cite{yamamoto16eng} used as training data around
200,000 pairs of an economic condition and its statement of the
reasons to learn a regression model so as to predict the business
sentiment of a given text. As a regression model, they used a
bidirectional Recurrent Neural Network with Long Short Term Memory
(LSTM)~\cite{hochreiter97:_long_short_term_memor}. Then, monthly
economic reports were fed to the learned model to compute a business
sentiment index. It is reported that the computed index was positively
correlated with both EWDI and Short‐term Economic Survey of Principal
Enterprise in Japan.


Aiba et al.~\cite{aiba18eng} used a similar model to compute a
business sentiment index from microblogs (tweets), and Kondo et
al.~\cite{kondo19eng} from bank's internal documents written from
interviews with their client corporations.  Goshima et
al.~\cite{goshima19eng} used a convolutional neural network and
Reuters news articles to compute a business sentiment index.






\section{S-APIR and its Application}
\label{sec:newspaper_nowcast}

As described in Section~\ref{sec:related_work}, Economy Watchers
Survey contains pairs of an economic condition and a statement of the
reasons. They are filled out manually by respondents and are quality,
valuable resources for machine learning. In the present work, we focus
on news articles as with Goshima et al.~\cite{goshima19eng} to compute
a business sentiment index named S-APIR.  However, in contrast to
Goshima et al. who fed news texts as they were to the learned model,
we attempt to filter out irrelevant news texts and to apply domain
adaptation as we will describe shortly in Section~\ref{sec:s-apir}.


Then, Section~\ref{sec:contribution} discusses an application of the
S-APIR index to temporally analyze any given factors that may/may not
influence business sentiment.  Business sentiment is formed
by many factors including monetary policies, stock prices, exchange
rates, unemployment rate, wages, overseas situations, etc.  However,
those factors do not equally influence business sentiment and it is
helpful for business economists if they could understand what factors
have a more/less influence to move up/down business sentiment in a
particular period. To this end, we propose a simple approach to
analyzing when and what factors contributed to S-APIR based on
predicted business sentiment.

\subsection{S-APIR Index}
\label{sec:s-apir}

This section describes three major components of our framework to
predict business sentiment from a news sentence.  The first component
is a regression model that takes a sentence and predicts the
sentiment of the input. The second is a classifier to filter out texts
irrelevant to the economy/business. Lastly, the third is domain adaptation
to update the parameters of an initial regression model to make them
more suitable for news texts.

\subsubsection{Regression Model}
\label{sec:regression}

For text classification and regression, it has been a common practice
to treat each word as an independent variable, where an input text is
represented as a Bag of Words (BoW) disregarding the
context~\cite{Manning:2008:IIR:1394399}.  However, it is desirable to
capture the differences of word meanings in different contexts and
word dependencies so as to properly represent the meaning of the text.

In recent years, RNN, combined with LSTM, has been popularly used to
represent text to consider the context.  This study also uses RNN but
with Gated Recurrent Unit
(GRU)~\cite{cho14:_proper_neural_machin_trans}, which can be seen as a
variant of LSTM.  Following the related work, we also use Economy
Watchers Survey to train the model, where the five-point-scale
economic conditions are converted to $\{-2, -1, 0, 1, 2\}$.


\subsubsection{Filtering}
\label{sec:filtering}

One could use a regression model to be described in the previous
section to predict business sentiment for any input text. However,
news texts we focus on in this study are in many genres which may be
irrelevant to the economy.  Using irrelevant sources would be harmful in
computing a business sentiment index.  Therefore, we attempt to filter
out such irrelevant news texts by treating them as outliers.

For this purpose, we adopt a one-class support vector machine
(SVM)~\cite{manevitz02:_one_svms_docum_class}.  In contrast to an
ordinal SVM used for binary classification, a one-class SVM can be
learned on documents in only one class and detect documents dissimilar
to the training documents as outliers.  We use Economy Watchers Survey
(statements of the reasons) as the training data for one-class SVM and
filter out news text dissimilar to the statements.

For text representation, one could use an output of an RNN or other
sentence
embeddings~\cite{cer18:_univer_senten_encod_englis,pagliardini18:_unsup_learn_senten_embed_compos_gram_featur}
so that the context could be better considered.  However, our
preliminary experiment showed that they resulted in detecting all news
texts as outliers.  The traditional BoW with term frequency-inverted
document frequency (tf-idf) term
weighting~\cite{Manning:2008:IIR:1394399} worked better for
this task and was used in this work.

\subsubsection{Domain Adaptation}
\label{sec:domain_adaptation}

The Economy Watchers Survey responses to be used for training a
GRU-RNN are different from news texts to be used for computing S-APIR
in terms of their writing styles, vocabularies, and expressions
(collectively called ``domains'').  Such differences between training
and testing would have a negative effect on the resulting performance
and adapting the domain of the learned model to the target domain
would benefit business sentiment prediction.

To this end, we explore domain adaptation by automatically creating
new training data from news texts. To be precise, we feed news
articles to an initial regression model (denoted as $M$) and predict
the business sentiment of each sentence. Then, we assume that
sentences with higher absolute sentiment scores would better represent
economic conditions either positively or negatively. We set predefined
positive and negative thresholds and treat the sentences with
higher/lower sentiment scores than the thresholds as positive and
negative examples, respectively. 

We use the training data to fine-tune the initial model $M$ to acquire
fine-tuned model $M'$.  More specific experimental settings (e.g.,
thresholds) are described in the evaluation in
Section~\ref{sec:evlaluation}.

\subsection{Temporal Analysis}
\label{sec:contribution}

Business sentiment is formed based on many factors including monetary
policies, tax reform, trade, and military conflicts.  This section
describes an approach to analyzing \textit{which} factor influenced
business sentiment \textit{when} and \textit{how much}.  Specifically,
we define the influence of word $w$ during time $t$, $p_{t,w}$, using
the predicted business sentiment.

We first assume that the sentiment $p_s$ of a sentence $s$ is the sum
of the sentiments of words ($w$) appearing in $s$ as follows:
\begin{equation}
  p_s=\sum_{w\in s} f_{s,w}\cdot p_{s,w} 
\end{equation}
where $f_{s,w}$ is the number of occurrences of word $w$ in $s$,
$p_{s,w}$ is the sentiment of $w$ in $s$. We further assume that all
the words, $w\in s$, equally influence the sentiment of $s$, that is,
\begin{equation}
  p_{s,w}=\frac{p_s}{|s|}
\end{equation}
where $|s|$ is the number of words composing $s$.  Here, let $S_t$
denote the set of news sentences published during $t$. Using $S_t$,
we define $p_{t,w}$ as the sum of $p_{s,w}$ over $S_t$, divided by the
number of sentences $|S_t|$.
\begin{equation}
  p_{t,w}=\frac{1}{|S_t|}\sum_{s\in S_t} f_{s,w}\cdot \frac{p_s}{|s|}
\end{equation}
Intuitively, S-APIR in time $t$ can be interpreted as the sum of the
influences of all the words appearing in texts published during
$t$.

\section{Evaluation}
\label{sec:evlaluation}

\subsection{Experimental Settings}
\label{sec:data}

For learning an RNN and a one-class SVM, we downloaded the Economy
Watchers Survey data from the web page of the Cabinet
Office\footnote{http://www5.cao.go.jp/keizai3/watcher/watcher\_menu.html}
in October 2018.  The number of the pairs of an economic condition and
a statement of the reasons was 216,741 in total, of which randomly
selected 90\% were used for training (and validation) and the rest
were used for testing.  Note that because Japanese text does not have
explicit word boundaries (such as spaces in the case of English), the
statements of the reasons were processed by a morphological analyzer,
MeCab\footnote{http://taku910.github.io/mecab/}, to be split into
words.

The parameters of a GRU-RNN were set as follows based on a preliminary
experiment on the Economy Watchers Survey data: the number of GRU
units per layer=512, the number of hidden layers=2, and the size of
the vocabulary=40,000.  Each word was represented as a word embedding
vector with 300 dimensions pretrained on
Wikipedia~\cite{bojanowski-etal-2017-enriching}.

To compute the S-APIR index, we used the titles and body texts of news
articles from the Nikkei Newspaper from 2013 to 2018. For domain
adaptation, we used separate Nikkei Newspaper published in 2010.  Each
article was split into sentences based on the Japanese period ``\Ja{。}'' and
each sentence was fed to the learned model to predict its sentiment.

\subsection{Evaluation on Economy Watchers Survey}
\label{sec:keiki_watcher_index}

First, we evaluated the learned (initial) GRU-RNN on the held-out test
data (10\% of Economy Watchers Survey).  Table~\ref{tab:regression}
compares our model and a ridge regression model in mean squared error
(MSE).  For ridge regression, we used the classic BoW representation
with tf-idf term weighting.  The MSE for GRU-RNN significantly
decreased as compared to that of ridge regression, which confirms that
our model predicted economic conditions more accurately than the ridge
regression.  This result is similar to the one reported in the related
work~\cite{yamamoto16eng}.

\begin{table}[htb]
  \small
  \centering
  \caption{Comparison between ridge regression and our GRU-RNN for
    predicting economic conditions.}
  \label{tab:regression}
  \smallskip
  \begin{tabular}{cc}
    \toprule 
    Model & MSE \\
    \midrule 
    Ridge regression & 0.509\\
    GRU-RNN & 0.351\\
    \bottomrule 
  \end{tabular}
\end{table}

\subsection{Evaluation of S-APIR}
\label{sec:nikkei_index}

This section compares our business sentiment index, S-APIR, and
existing business sentiment index, namely EWDI.  It should be
emphasized, however, that S-APIR is not intended to replace EWDI, but
rather to be a new index using newspaper as the source of information.
There is no ground truth for a business sentiment index and EWDI is
also an index calculated based on the limited number of 2,050
respondents.  The purposes of the comparison in this section are (1)
to ensure that S-APIR has a similar trend to the existing index and
(2) to investigate the characteristics of S-APIR when the two indices
diverge.

\subsubsection{Results}
\label{sec:results}

\begin{figure}[tb]
  \centering
  \includegraphics[width=\linewidth]{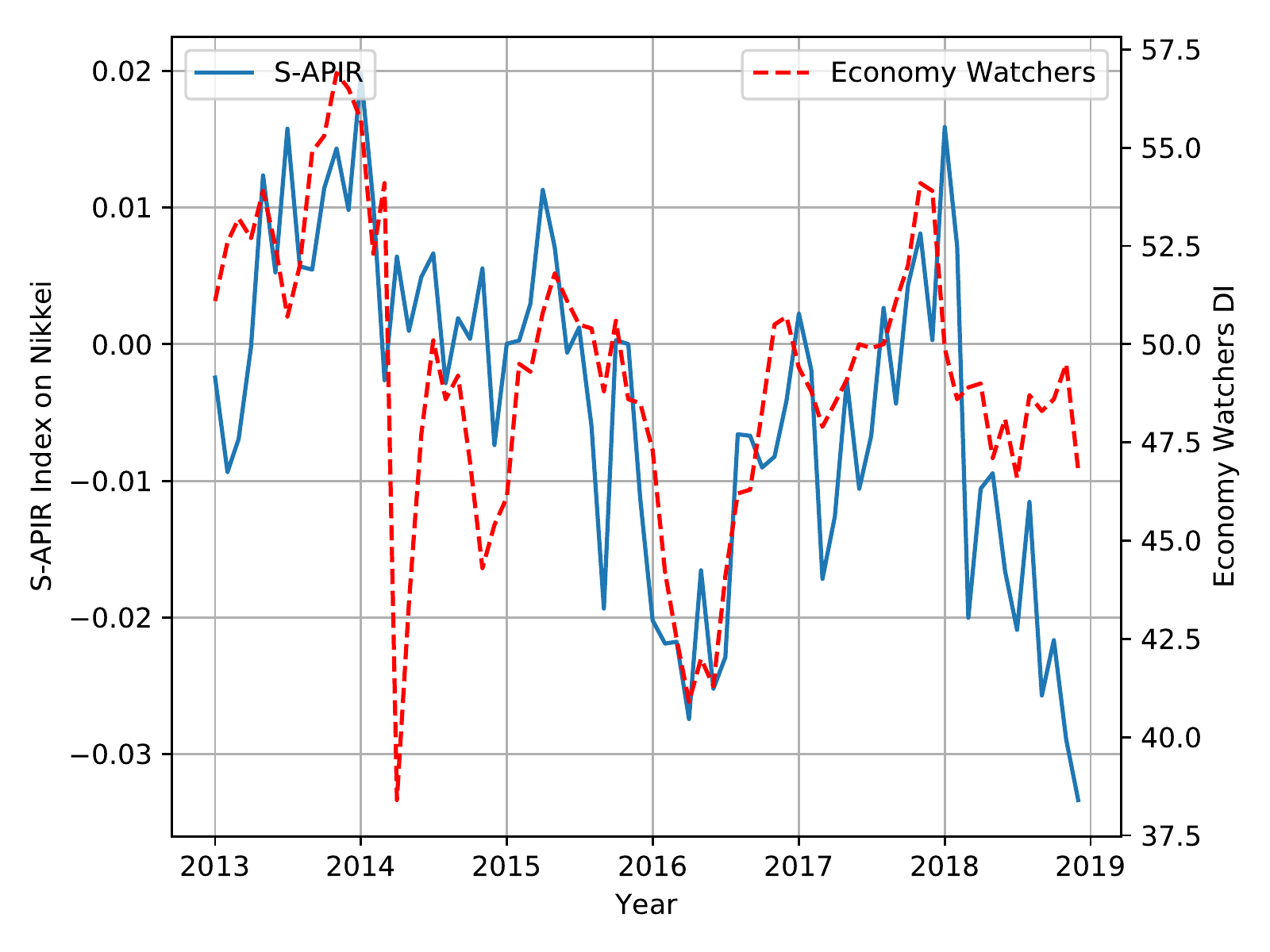}
  \caption{Comparison between S-APIR and EWDI ($r=0.546$).}
  \label{fig:nikkei}
\end{figure}

Using the initial GRU-RNN learned as described in
Section~\ref{sec:keiki_watcher_index}, we first computed S-APIR on
Nikkei Newspaper from 2013 to 2018.  Note that as business sentiment
is predicted for each sentence, they were aggregated monthly by taking
the average to compute S-APIR for each month.  Figure~\ref{fig:nikkei}
shows the computed S-APIR and EWDI for comparison.  We can observe
that the two indices show roughly similar movements.  In effect, they
were found to be positively correlated ($r=0.546$).


\begin{figure}[tb]
  \centering
  \includegraphics[width=\linewidth]{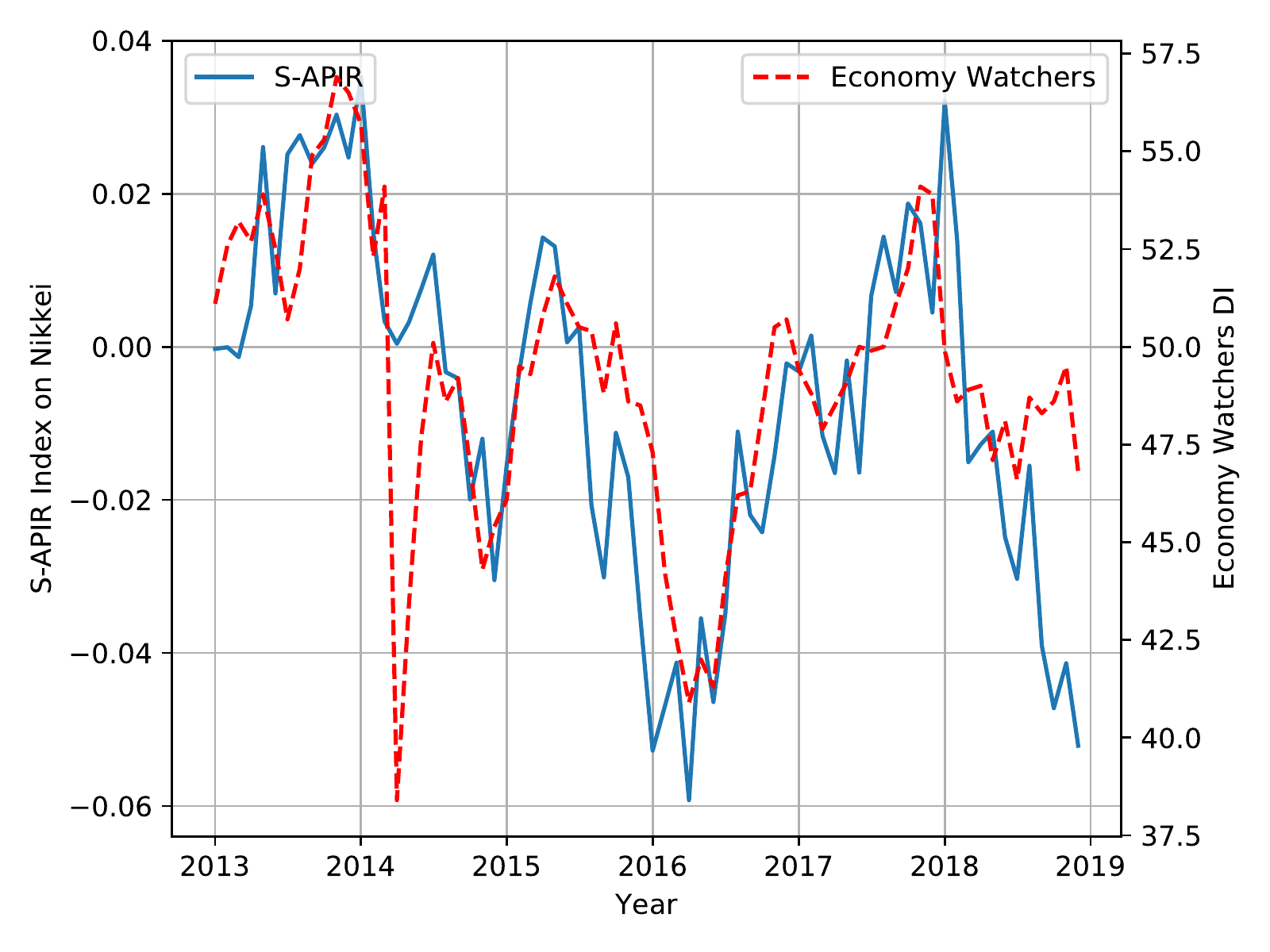}
  \caption{Comparison between S-APIR and EWDI after filtering
    ($r=0.686$).}
  \label{fig:nikkei_filtered}
\end{figure}

Next, we applied the one-class SVM to detect and filter out outliers
and recomputed S-APIR by using only the texts relevant to the economy.
Figure~\ref{fig:nikkei_filtered} shows the result.  Overall, S-APIR
exhibited a more similar trend to EWDI and their correlation
coefficient significantly increased from 0.546 to 0.686.  The result
confirms the effectiveness of the filtering process by the one-class
SVM.

As shown above, the GRU-RNN learned on Economy Watchers Survey
(statements of the reasons) can be used to compute the business
sentiment index based on news texts.  However, statements of the
reasons and news texts have different characteristics and the model
learned on the former may not be suitable for the latter.  Thus, we
applied domain adaptation as described in
Section~\ref{sec:domain_adaptation}.  To be specific, we took the
following procedure:

\begin{enumerate}
\item Extracted titles and body texts of news articles from Nikkei
  Newspaper 2010 and split them into sentences and then into words.
\item Applied the one-class SVM and filtered out outliers.
\item Applied the GRU-RNN to predict the sentiment of each sentence.
\item Identified the sentences with sentiment scores greater (lower)
  than a predefined threshold $t_{high}$ ($t_{low}$).  To determine the
  thresholds, we looked at the histogram of the sentiment scores and
  experimentally set $t_{high}=0.8$ and $t_{low}=-1.0$.  As a result,
  we obtained 18,947 positive instances and 10,868 negative instances.
  We gave the former ``2'' as their labels, and the latter ``$-2$''.
\item Fed the generated training data to the learned (initial) GRU-RNN
  for fine-tuning.
\end{enumerate}

\begin{figure}[tb]
  \centering
  \includegraphics[width=\linewidth]{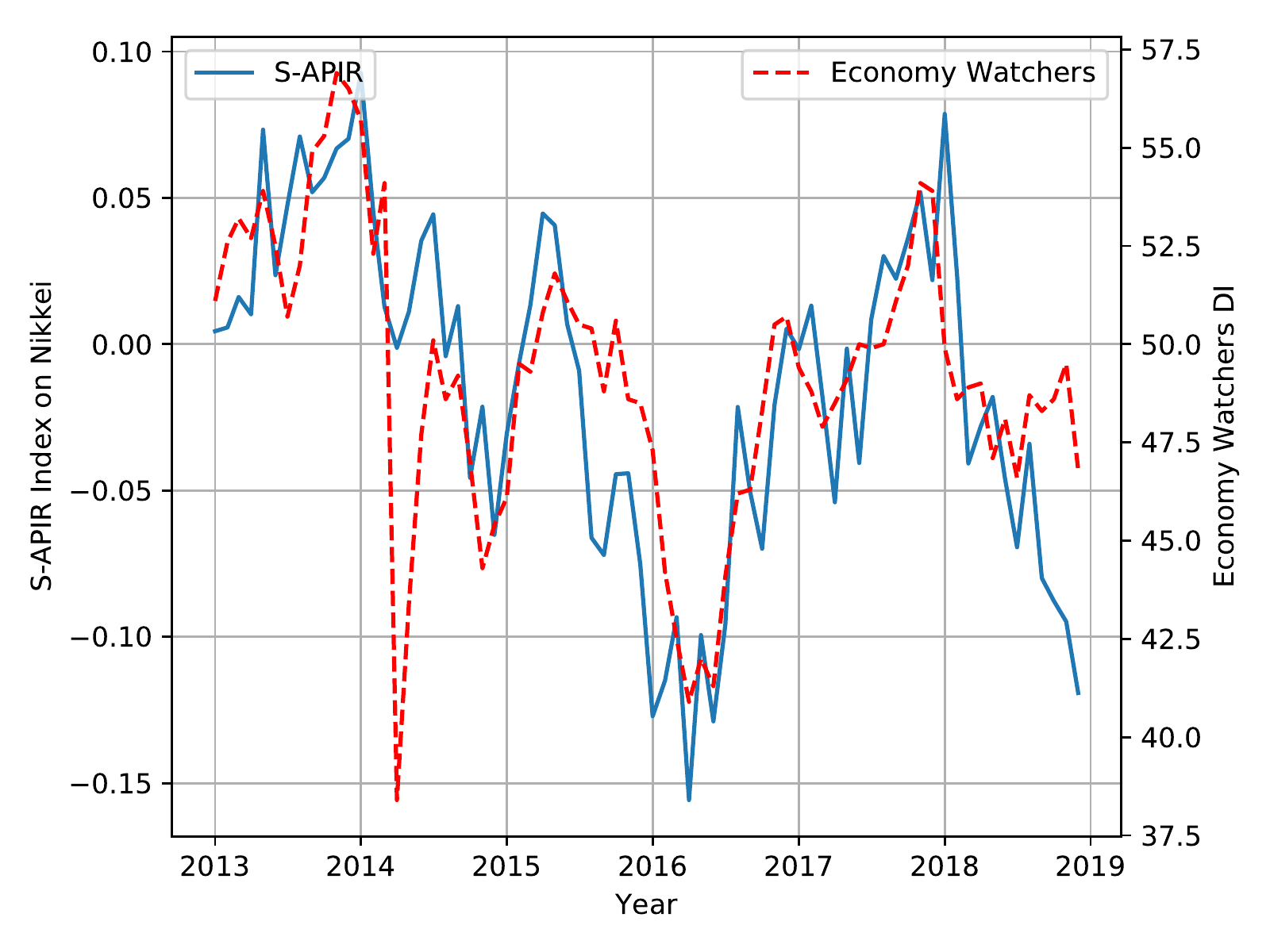}
  \caption{Comparison between S-APIR and EWDI after domain adaptation
    ($r=0.701$).}
  \label{fig:da1}
\end{figure}

Then, we used the fine-tuned model to recompute S-APIR after
filtering.  The result is shown in Figure~\ref{fig:da1}.  The
correlation coefficient marginally increased to 0.701.

Note that EWDI dropped sharply in April 2014, where there is a large
deviation from the S-APIR index.  This is when sales tax was increased
from 5\% to 8\% in Japan and it is interesting to learn that the S-APIR
index is much less affected by the tax increase.  We conjectured that
this might be due to the fact that 70\% of the respondents of Economy
Watchers Survey had occupations related to households.  Therefore,
factors that have more influence on households (e.g., tax increase) may
have more influence on EWDI as well.  To verify the intuition, we
compared S-APIR and a variant of EWDI computed based on responses only
from those who have occupations related to industries.  As a result,
the correlation coefficient indeed increased from 0.701 to 0.819.
This result suggests that the S-APIR index, computed from Nikkei
Newspaper, reflects business sentiment in industries more strongly.

\subsubsection{Effect of Domain Adaptation}
\label{sec:domain_adaptation_result}

The previous section empirically showed that domain adaptation
increased the correlation between S-APIR and EWDI but only marginally.
Here, we look into the models before/after domain adaptation to
investigate how the model changed.  To this end, we fed single words
as inputs to each model and predict the sentiments of the words.

Table~\ref{tab:positive} compares ten words with higher sentiment for
each model on descending order of the sentiment scores, where words
with ``$\uparrow$'' indicate those went up after domain adaptation and
those with ``$\downarrow$'' indicate went down.  While the rankings of
``\Ja{好調} (good condition)'', ``\Ja{享受} (to enjoy)'', ``\Ja{回復} (recovery)'',
and others went up, those of ``\Ja{最高} (best)'', ``\Ja{絶好調} (best
condition)'', and ``\Ja{伸} (to grow)'' went down.

\begin{table*}[htb]
  \small
  \centering
  \caption{Sentiment of top 10 words before/after domain adaptation.}
  \label{tab:positive}
  \smallskip
  \begin{tabular}{lc|clc}
    \hline 
    \multicolumn{2}{c|}{Before} & \multicolumn{3}{c}{After} \\
    \hline 
\Ja{快調} (good condition)    & 0.738 & $\uparrow$  & \Ja{好調} (good condition)   & 1.591\\ 
\Ja{絶好調} (best condition) & 0.688 & $\uparrow$   &\Ja{享受} (to enjoy)   & 1.503\\ 
\Ja{好調} (good condition) & 0.671 & $\uparrow$  & \Ja{復調} (recovery)   & 1.502\\ 
\Ja{最高} (best)   & 0.659 & $\uparrow$  & \Ja{好転} (to improve) & 1.488\\ 
\Ja{上々} (excellent)   & 0.654 & $\uparrow$  &  \Ja{向上} (to improve)   & 1.485\\ 
\Ja{好転} (to improve)   & 0.654 & $\downarrow$ & \Ja{最高} (best)   & 1.484\\ 
\Ja{伸} (to grow)      & 0.632 & $\uparrow$  & \Ja{刺激} (stimulation)   & 1.469\\ 
\Ja{売れれ} (to sell)  & 0.630 & $\downarrow$ & \Ja{絶好調} (best condition)  & 1.456\\ 
\Ja{上向い} (to look up) & 0.624 & $\downarrow$ & \Ja{伸} (to grow)      & 1.449\\ 
\Ja{着実} (steady)   & 0.617 & $\uparrow$  & \Ja{図り} (to plan)   & 1.437\\
    \hline 
  \end{tabular}
\end{table*}

Similarly, Table~\ref{tab:negative} compares ten words with lower
sentiment scores. While ``\Ja{悪化} (to worsen)'', ``\Ja{激減} (marked
decrease)'', and others went up in the list, ``\Ja{舗装} (pavement)'' and
``\Ja{壊滅} (complete destruction)'' went down.  In both cases, we can
observe that words that would be more often used in news text went up
and those used more often in survey responses went down after domain
adaptation.

\begin{table*}[htb]
  \small
  \centering
  \caption{Sentiment of top 10 words before/after domain adaptation.}
  \label{tab:negative}
  \smallskip
  \begin{tabular}{lc|clc}
    \hline 
    \multicolumn{2}{c|}{Before} & \multicolumn{3}{c}{After} \\
    \hline 
    \Ja{舗装} (pavement)  & $-$1.543 & $\uparrow$ &             \Ja{悪化} (deterioration)    & $-$2.029\\ 
    \Ja{不通} (blocked)  & $-$1.441 & $\uparrow$ &             \Ja{激減} (marked decrease)   & $-$2.028\\
    \Ja{最悪} (worst)  & $-$1.355 &   --     &       \Ja{最悪} (worst)   & $-$2.026\\
    \Ja{壊滅} (complete destruction)   & $-$1.334 &   $\downarrow$   &         \Ja{舗装} (pavement)    & $-$2.026\\
    \Ja{急変} (sudden change)  & $-$1.303 &    --      &     \Ja{急変} (sudden change)   & $-$2.019\\
    \Ja{全滅} (complete destruction)  & $-$1.285 & $\downarrow$    &          \Ja{壊滅}
                                              (complete destruction)   & $-$2.018\\
    \Ja{追い打ち} (aggravation)       & $-$1.282 &   $\uparrow$  &  \Ja{低落} (decline)  & $-$2.017\\
    \Ja{激減} (marked decrease)   & $-$1.278 &    $\uparrow$     &      \Ja{減益} (profit fall)  & $-$2.017\\
    \Ja{エスカレート} (to escalate)   & $-$1.257 &   $\uparrow$  &  \Ja{損失} (loss)   & $-$2.013\\
    \Ja{漁船} (fishing vessel)  & $-$1.232 &   $\uparrow$        &    \Ja{急落}
                                        (significant fall)    & $-$2.011\\
    \hline 
  \end{tabular}
\end{table*}

Notice that while we predicted business sentiments of individual words
to investigate the changes of the model after domain adaptation,
resulting business sentiment scores can be seen as business sentiment
polarities of the words.  That is, the results can be used as
a sentiment dictionary in the economic/financial domain.  There are
several sentiment dictionaries in the general domain but only a few in
the economic/financial domain and, as far as we know, none exists in
Japanese.  Therefore, the result can be useful resources for business
sentiment analysis.

\subsection{Temporal Analysis of Words}
\label{sec:contribution-result}

\begin{figure}[htb]
  \centering
  \includegraphics[width=\linewidth]{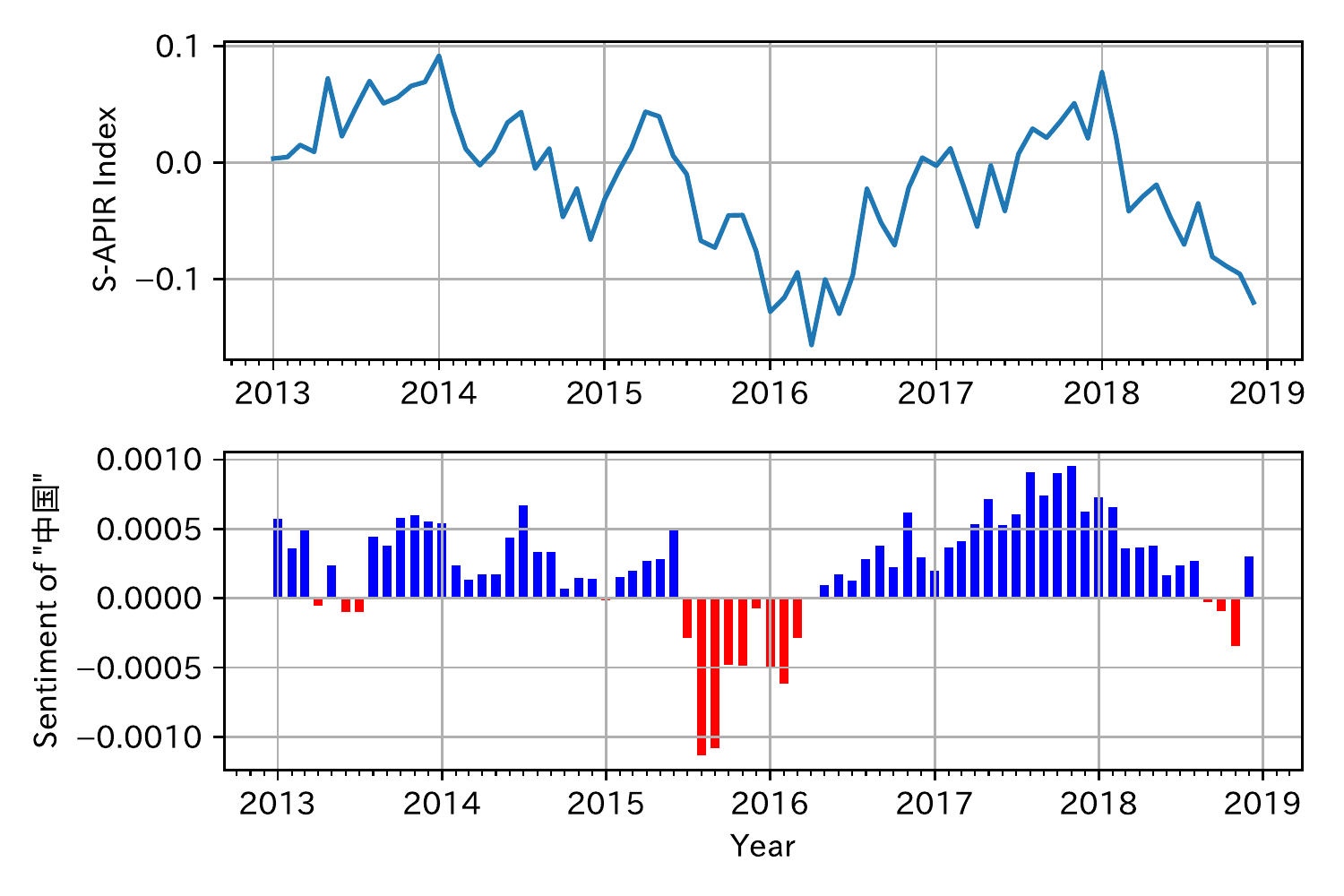}
  \caption{Temporal influence of ``\protect\Ja{中国} (China)'' on S-APIR.}
  \label{fig:china}
\end{figure}

\begin{figure}[htb]
  \centering
  \includegraphics[width=\linewidth]{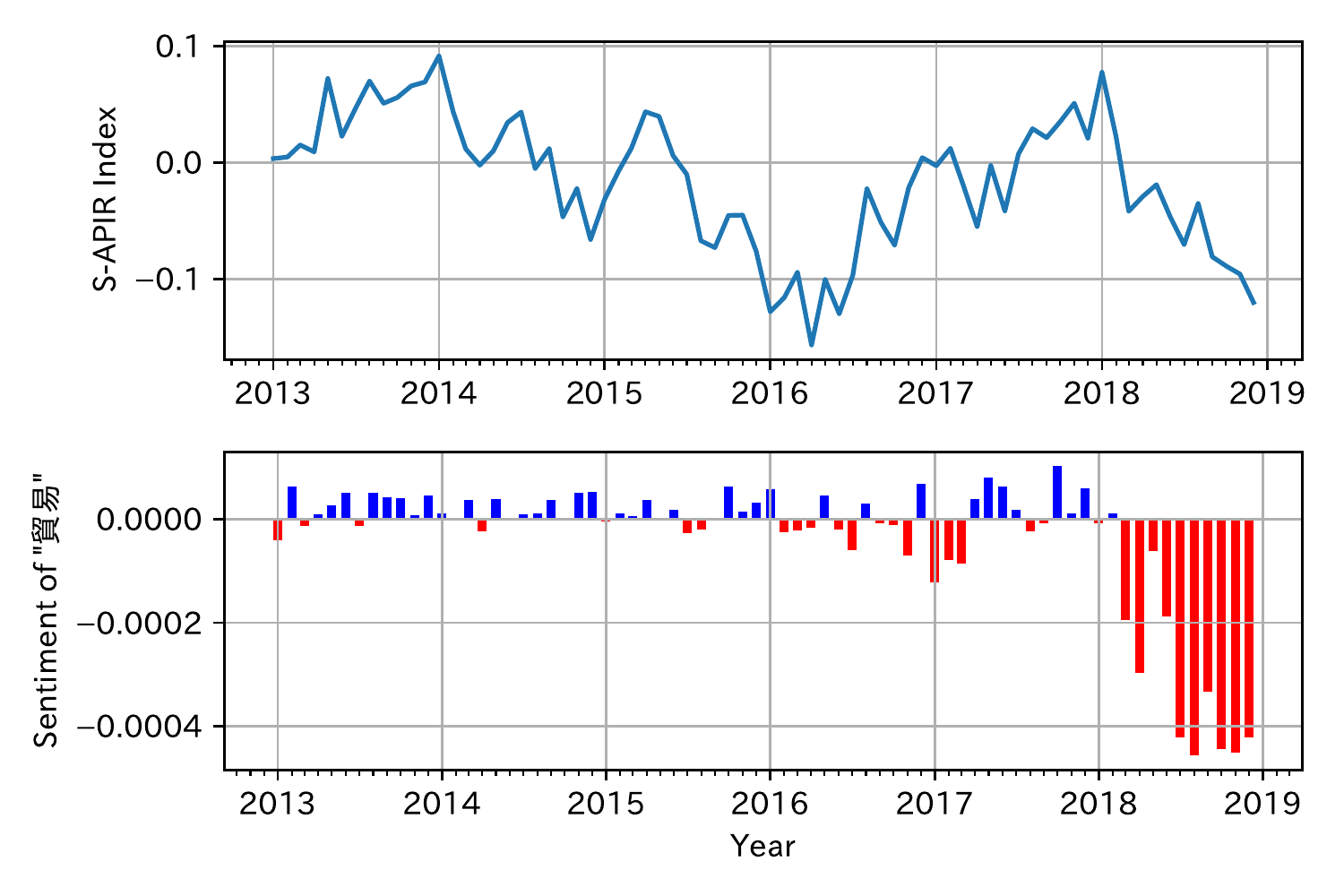}
  \caption{Temporal influence of ``\protect\Ja{貿易} (trade)'' on S-APIR.}
  \label{fig:trade}
\end{figure}

Lastly, we temporally analyzed the influence of a given factor (word)
on S-APIR as described in Section~\ref{sec:contribution}.  While our
proposed approach can analyze any given words, here we focused
on a few representative examples.  Specifically, we computed the
influence of ``\Ja{中国} (China)'' and ``\Ja{貿易} (trade)'' for example.  The
results are shown in Figure~\ref{fig:china} and
Figure~\ref{fig:trade}, respectively.  In both figures, the upper
graph is S-APIR from Figure~\ref{fig:da1} and is shown for reference.

In Figure~\ref{fig:china}, S-APIR and the influence of China generally
have similar movements and thus situations about China appear to be
one of the major factors influencing the business sentiment
index. Especially, from the middle of 2015 to the beginning of 2016,
the influence of China is strongly negative, which is pushing down the
business sentiment in Japan.  This period is a time when the Chinese
economy deteriorated rapidly due to the crash of China's stock market.

Then, looking into Figure~\ref{fig:trade}, we can observe that
``trade'' has not had much influence from 2013 to the beginning of
2018, whereas the situation has changed thereafter and started to show
a strong negative influence on business sentiment.  This reflects the
US-China trade dispute that began in late 2018.

\section{Conclusions}
\label{sec:conclusions}

This paper reported on our ongoing work to develop a new business
sentiment index, called S-APIR, based on news texts and to use the
index to temporally analyze the factors that influence business
sentiment.  We used a one-class SVM to identify news texts related to
the economy and fed them to a GRU-RNN regression model to predict the
business sentiment of input news text.  The GRU-RNN was initially
trained on Economy Watchers Survey and then fine-tuned on news texts
for domain adaptation.  Through our evaluation using Nikkei Newspaper
articles, it was demonstrated that S-APIR has a positive correlation
with an existing business sentiment index and that the correlation
becomes even higher when compared to a variant of the index related to
industries.  This result indicates that S-APIR is a business
sentiment index reflecting that of industries more strongly.
Moreover, by dividing sentence sentiment into word sentiments and
summing over sentences, it was shown that any given factor that
may/may not have an influence on business sentiment can be temporally
analyzed.


\section*{Acknowledgments}

This work was conducted partly as a research project ``development and
application of new business sentiment index based on textual data'' at
APIR and was partially supported by JSPS KAKENHI \#JP18K11558 and
MEXT, Japan. We thank Hideo Miyahara, Hiroshi Iwano, Yuzo Honda,
Yoshihisa Inada, Yoichi Matsubayashi, and Akira Nakayama for their
support.  The Nikkei data were provided by APIR.

\bibliographystyle{named}
\bibliography{../reference}

\end{document}